**FinVision: Multimodal AI for Visualizing Financial Narratives** [*]

Yixuan Tang[a], Hui Wang[b], Yi Yang[c], Allen H. Huang[d]

**November 2026**

**Abstract**

Financial narratives contain rich information about firms' performance and business conditions. Conventional text analysis often summarizes this information into numerical representations, which may not fully preserve the dynamic structure described in the narrative. In this study, we develop FinVision, a domain-specific multimodal generative model that translates financial narratives into visual representations of contemporaneous market dynamics. Using 87,940 paired observations of Management Discussion and Analysis (MD&A) disclosures and stock price charts, FinVision learns the average relationship between a firm's narrative account of a fiscal year and its stock-price dynamics over the same period. Given an MD&A alone, the trained model generates a text-implied stock chart, extending financial text analysis beyond numerical representations toward text-to-visual generation. We validate FinVision in two ways. First, held-out tests show that the model generates stock-chart representations that closely resemble observed charts and outperform direct LLM prompting, general-purpose text-to-image models, and the same architecture without financial-domain adaptation. Second, we examine the economic validity of the generated representations by comparing text-implied and observed stock charts. The resulting discrepancy is systematically associated with weaker investor responses around subsequent 10-K filings and with worse subsequent market and operating outcomes. Overall, our findings show how domain-adapted multimodal generative AI can create a new visual representation of financial text and provide an additional modality for studying business communication.

---

[*] We gratefully acknowledge the financial support of the grant (WEB25BM02).
[a] *Department of Information System, Business Statistics and Operation Management, HKUST* (ytangch@connect.ust.hk)
[b] *Department of Accounting, Renmin University of China* (wanghui@rmbs.ruc.edu.cn)
[c] *Department of Information System, Business Statistics and Operation Management, HKUST* (imyiyang@ust.hk)
[d] *Department of Accounting, HKUST* (allen.huang@ust.hk)

## 1. Introduction

*"In the first quarter of 2025, we saw high demand on exports out of Asia and continued to see high demand on exports out of South Asia in the second quarter 2025… However, softening demand and additional available capacity for ocean freight resulted in declines in ocean sell and buy rates starting in the second quarter."*

*— Expeditors International, 2025 Form 10-K*

Financial narratives are an important form of business communication. Firms use annual reports, earnings announcements, conference calls, and other narrative disclosures to describe how their businesses evolve over time. The opening example illustrates this dynamic nature: the same annual report can describe stronger activity in the first half of the year and softer conditions later in the year.

The growing availability of financial narratives has motivated extensive research using natural language processing (NLP) and, more recently, large language models (LLMs) to extract information from these texts. A common approach is to transform financial narratives into summary representations, such as sentiment, readability, risk, or low-dimensional embeddings, and then use these representations in downstream prediction or empirical analysis (Li, 2008, 2010; Loughran and McDonald, 2011; Chen, Kelly, and Xiu, 2022; Huang, Wang, and Yang, 2023; de Kok, 2025; Siano, 2025). These approaches have substantially expanded our ability to analyze business communication at scale. At the same time, because they typically summarize a long narrative into a small number of variables or vectors, they may not fully preserve the dynamic structure contained in the text. In the opening example, for instance, a dictionary-based approach might assign a broadly neutral sentiment score to the entire report, even though the narrative describes a more nuanced pattern of stronger conditions early in the year and weaker conditions later.

In this study, we explore a different representational paradigm: translating financial narratives into visual charts. Our empirical setting focuses on the MD&A section of 10-K filings and the firm's stock price chart over the same fiscal year. The two provide related views of the same firm-period economic environment: the MD&A summarizes management's account of the developments, risks, and conditions experienced during the year, while the stock chart provides a structured representation of how the market evolved over the same period. Prior research also shows that stock price charts contain economically meaningful information (Jiang, Kelly, and Xiu, 2023).

Figure 1 illustrates how our approach generalizes empirical analysis of financial text. Traditional studies often transform a financial narrative into a researcher-defined construct, such as sentiment, and use a regression model to estimate its average relationship with a scalar market outcome, such as annual stock return. We retain the same underlying empirical logic of learning an average relationship from many observations, but generalize the framework along three dimensions. First, the input moves from a pre-specified textual construct to the financial narrative itself. Second, the outcome moves from a scalar variable to a structured visual representation of market dynamics. Third, the mapping function moves from a regression model to a nonlinear multimodal generative model. In this sense, the framework generalizes the conventional text-to-variable setting into a text-to-visual setting.

The multimodal model therefore learns, on average, what visual market patterns are associated with different financial narratives. We do not assume that a narrative uniquely determines the realized stock-price path, just as a single explanatory variable in a regression does not fully determine the observed outcome. Stock prices reflect many other sources of information beyond the disclosure. The objective is instead to learn the systematic average mapping between financial narratives and contemporaneous market dynamics. Differences between the text-implied and realized stock charts can then be viewed as residual variation around this learned relationship and may themselves contain useful information.

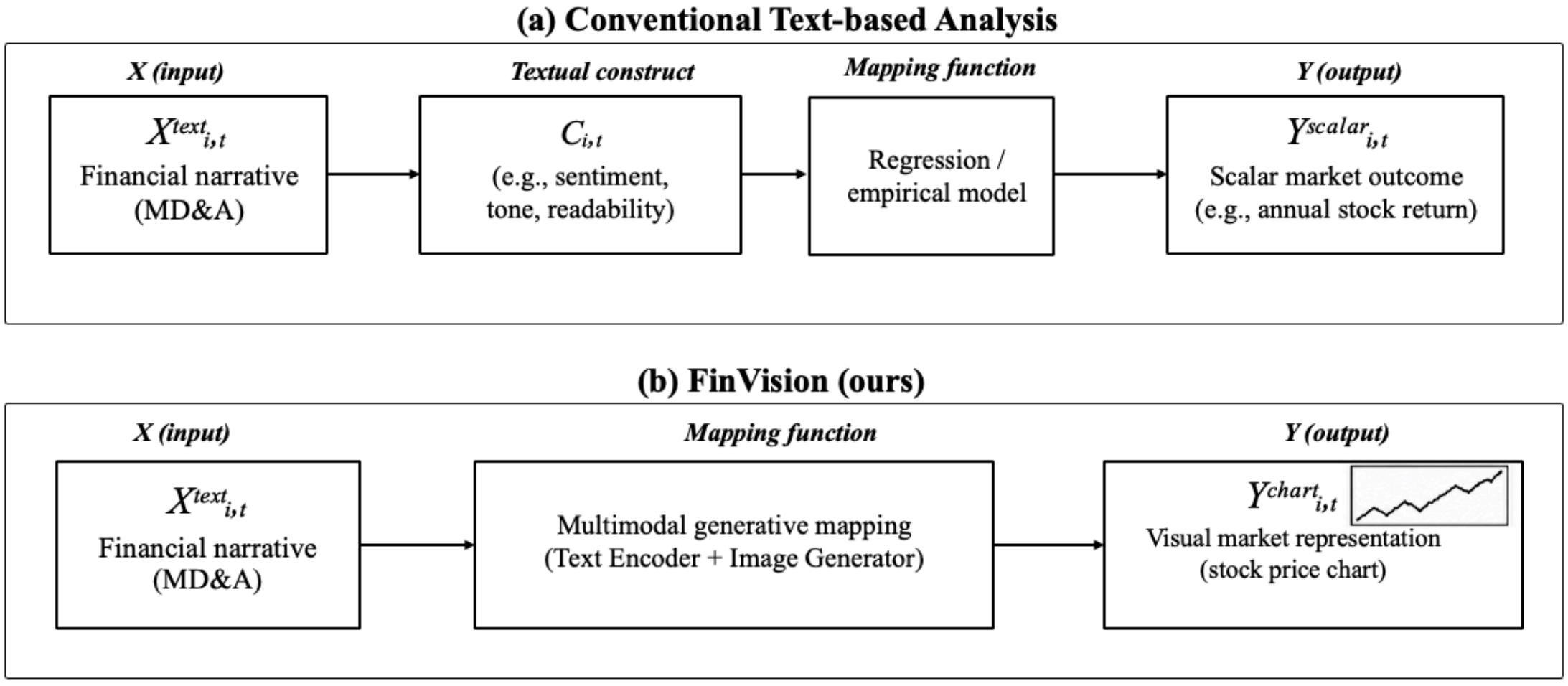

**Figure 1**: A conceptual comparison between FinVision and conventional text-based analysis. From text–scalar regression to text–visual generative mapping.

Recent advances in multimodal AI create a new opportunity to study this possibility by enabling models to learn relationships across heterogeneous forms of information and generate content in one modality conditional on another. In particular, general-purpose text-to-image models have achieved remarkable success in generating images that closely reflect natural-language descriptions. However, applying these models to financial narratives is not sufficient. Stock-price charts are structured outputs, and the relationship between disclosure language and market-chart patterns is domain specific. This makes financial-domain adaptation central to the task: the model must learn the mapping from narrative disclosures to market-chart structure from paired financial data. More fundamentally, it remains unclear whether multimodal AI can extract the dynamic information contained in long financial narratives and translate it into economically meaningful visual patterns.

To address these challenges, in this study, we propose **FinVision**, a domain-specific multimodal generative model that translates financial narratives into stock price charts.

FinVision employs an encoder-decoder architecture that combines a vision-language model for processing long financial narratives, an image autoencoder for representing financial charts, and a diffusion transformer for conditional image generation. During training, the model learns from paired MD&A disclosures and contemporaneous stock charts, while the text encoder and image autoencoder remain fixed. After the model is trained, given an MD&A as input, FinVision generates a *text-implied stock price chart*: a visual representation of the market dynamics associated with the information conveyed in the narrative. In our empirical setting, we construct 87,940 pairs of Management Discussion and Analysis (MD&A) disclosures from 10-K filings and stock price charts for the corresponding firm-fiscal years from 1996 to 2020. Although the underlying model components are established multimodal architectures, their integration and financial-domain adaptation enable a new text-to-visual analysis of corporate disclosure.

We validate FinVision in two complementary ways. First, we conduct visual validation on an out-of-sample of firm-years that are not used in model training. Given only the MD&A narrative, FinVision generates a text-implied stock chart, which we compare with the contemporaneous observed chart. Across multiple image-similarity metrics, FinVision produces representations that are substantially closer to the observed charts than those generated by direct LLM prompting, general-purpose text-to-image models, and a version of the same FinVision architecture trained on general web images.

Second, we examine the economic validity of the generated visual charts. Because FinVision learns the average relationship between financial narratives and contemporaneous market dynamics from a large number of paired observations, the generated chart provides a text-implied benchmark for the market pattern associated with a given narrative. We compare this text-implied chart with the firm's observed chart and use their visual difference as a residual-like measure of narrative–market discrepancy. We then examine whether this difference is systematically related to economically meaningful outcomes. Greater discrepancy is associated with weaker investor reactions around subsequent 10-K filings and with worse subsequent market and operating performance. Moreover, these economic associations are substantially weaker or absent when discrepancy is constructed using general-purpose image-generation models or FinVision without financial-domain adaptation. These findings suggest that the difference between generated and observed visual representations contains economically relevant information and illustrate how multimodal generation can support empirical analysis beyond image generation itself.

Our study makes two contributions. First, we make a methodological contribution by introducing a new paradigm for financial text analysis. Conventional approaches typically transform narratives into variables that are then related to numerical outcomes. FinVision instead learns a financial domain-adapted multimodal mapping from financial narratives directly to structured visual representations of market dynamics. To the best of our knowledge, this is the first attempt in financial communication to generate stock-chart representations directly from narrative disclosures. Therefore, the methodological novelty lies in adapting existing multimodal generative architectures to the financial domain to create a new visual representation space for analyzing financial text.

Second, we introduce a new visual modality for empirical analysis of financial communication. By expressing narrative information in the same visual space as observed market dynamics, FinVision enables direct comparison between text-implied and realized market patterns. We demonstrate this possibility by constructing narrative–market discrepancy from the visual distance between the two representations and showing that it is systematically associated with investor responses and future firm outcomes. This creates a new analytical dimension for studying corporate disclosure and shows how multimodal generative AI can expand empirical research beyond conventional text-based variables.

## 2. Literature Review

### 2.1 Computational Analysis of Financial Narratives

Financial narratives contain rich information about firms' performance, operating conditions, risks, and future prospects. A large literature in accounting, finance, and information systems has therefore developed computational methods to extract information from corporate disclosures and other financial texts. Early studies primarily relied on dictionaries and hand-crafted linguistic features to quantify characteristics such as readability, tone, forward-looking content, and textual similarity (Li, 2008, 2010; Loughran and McDonald, 2011, 2014; Brown and Tucker, 2011). These measures transformed otherwise unstructured narratives into numerical variables that could be incorporated into conventional empirical analyses.

Advances in natural language processing and machine learning subsequently enabled researchers to learn textual representations directly from financial data. Ke, Kelly, and Xiu (2019), for example, develop a return-predictive sentiment measure from news text, illustrating how textual representations can be learned based on their relationship with market outcomes. More recently, pretrained language models and large language models have substantially improved the ability to capture contextual and domain-specific information in financial text (Chen, Kelly, and Xiu, 2022; Huang, Wang, and Yang, 2023; de Kok, 2025; Siano, 2025). These approaches can learn representations tailored to particular analytical objectives, including the prediction of financial outcomes.

Despite substantial methodological progress, most existing approaches share a common representational structure. Financial narratives are typically mapped into textual features, latent vectors, discrete labels, or numerical outcomes. Such representations are highly useful for particular constructs, such as sentiment, or prediction targets, such as stock return, but they generally do not explicitly preserve how business conditions evolve over the reporting period. Our study builds on this literature but considers a different modality: visual representations. We examine whether financial narratives can be translated into structured visual representations of corresponding market dynamics that explicitly capture the temporal patterns described in the text.

### 2.2 Visual and Multimodal Analysis of Financial Information

Financial information is inherently multimodal, and a growing literature has examined visual representations of financial information. One stream represents numerical market data as images and applies computer vision methods to financial prediction. Jiang, Kelly, and Xiu (2023), for example, transform historical stock data into price charts and use convolutional neural networks to predict future stock returns. This literature establishes financial charts as a useful representation of market information and shows that visual patterns can support prediction and other analytical tasks. A second stream incorporates visual information together with text and other modalities. Studies of corporate disclosures have examined visual information in annual reports and other financial documents alongside their textual content (Christensen, Fronk, Lee, and Nelson, 2024; Basu, Ji, and Luo, 2025; Ben-Rephael et al., 2026; Deng et al., 2026). Related research combines earnings-call transcripts with managers' vocal characteristics and other non-textual signals (Baik, Kim, Kim, and Yoon, 2025; Ewertz, Knickrehm, Nienhaus, and Reichmann, 2026). Collectively, this literature shows that non-textual modalities can provide information complementary to financial narratives and can enrich the analysis of corporate communication and financial outcomes.

Recent financial vision-language research further integrates textual and visual information through multimodal models (Bhatia et al., 2024; Luo et al., 2025). Financial question-answering studies train models to interpret charts and answer questions about their contents, while financial document-understanding models jointly process text, tables, figures, and document layouts. These studies extend financial NLP from text-only analysis to multimodal understanding and demonstrate the ability of vision-language models to reason over visually represented financial information.

However, existing visual and multimodal financial research has predominantly used visual information, not the financial narrative, as input. Image-based models use observed charts to predict outcomes, while vision-language models use charts to generate textual answers. Thus, less attention has been paid to generating financial visual representations directly from text. Our study considers the opposite direction: we take financial narratives as input and generate the corresponding visual representation of market dynamics.

### 2.3 Text-to-Visual Generation

Recent advances in multimodal AI provide the technical foundation for moving from visual understanding to visual generation. Early text-to-image models established the feasibility of generating images conditioned on natural-language descriptions, while subsequent diffusion-based approaches substantially improved image quality and text-image alignment (Rombach et al., 2022). More recent architectures based on diffusion transformers further improve the scalability and quality of text-conditioned image generation (Peebles and Xie, 2023). Together, these developments have substantially expanded the ability of generative models to translate linguistic information into visual representations.

Most text-to-image research has focused on general-purpose image generation, where models translate natural-language descriptions into realistic visual content. More recently, researchers have extended these methods to specialized domains in which domain-specific text can be mapped into structured visual outputs. In medicine, for example, generative models have been used to synthesize radiological images from textual descriptions or clinical information (Bluethgen et al., 2025). These studies suggest that cross-modal generation can capture meaningful relationships between specialized language and visual representations of the same underlying phenomenon.

However, this generative direction has received limited attention in financial research. Existing financial multimodal models largely focus on interpreting charts or tables and producing textual outputs, rather than generating financial visual representations from textual inputs. At the same time, the broader text-to-image literature has primarily studied generated images as the target output, with limited attention to their use as representations that can subsequently support financial measurement and empirical analysis.

### 2.4 Research Gap

The streams of research reveal an important gap. First, existing computational research on financial communication primarily transforms text into textual features, or numerical outcomes, with limited attention to visual representations as an output. Second, research on visual and multimodal financial information has demonstrated the value of financial charts and other visual data, but predominantly uses visual information as an input for prediction or generates textual outputs from visual inputs. Third, recent multimodal AI research has established powerful text-to-image capabilities, but has largely focused on general-domain visual generation. It remains unclear whether multimodal AI can extract the dynamic information contained in long financial narratives and translate it into economically meaningful visual patterns. To the best of our knowledge, we are among the first to develop a text-to-visual generative approach that translates financial narratives into visual representations of corresponding market dynamics. We empirically show that FinVision can generate visually meaningful stock-price charts from financial text and that the resulting representations contain economically relevant information. This extends financial text analysis from numerical representations to structured visual representations and offers a new analytical way to encode meaningful information from financial narratives.

## 3. FinVision

We propose FinVision, a domain-specific multimodal generative model that translates financial narratives into visual representations of contemporaneous market dynamics. We pair MD&A narratives with stock price charts from the same firm and fiscal year as the training input (Section 3.1). We represent stock prices as standardized visual charts that preserve their temporal structure (Section 3.2). FinVision combines a text encoder, an image autoencoder, and a Diffusion Transformer to learn the cross-modal relationship between financial narratives and stock-chart representations (Section 3.3). We then train FinVision using the paired narratives-charts (Section 3.4). After FinVision is trained, given a new MD&A as input, the trained model generates a text-implied stock chart representing the market dynamics associated with the information conveyed in the narrative (Section 3.5). Figure 2 provides an overview of FinVision.

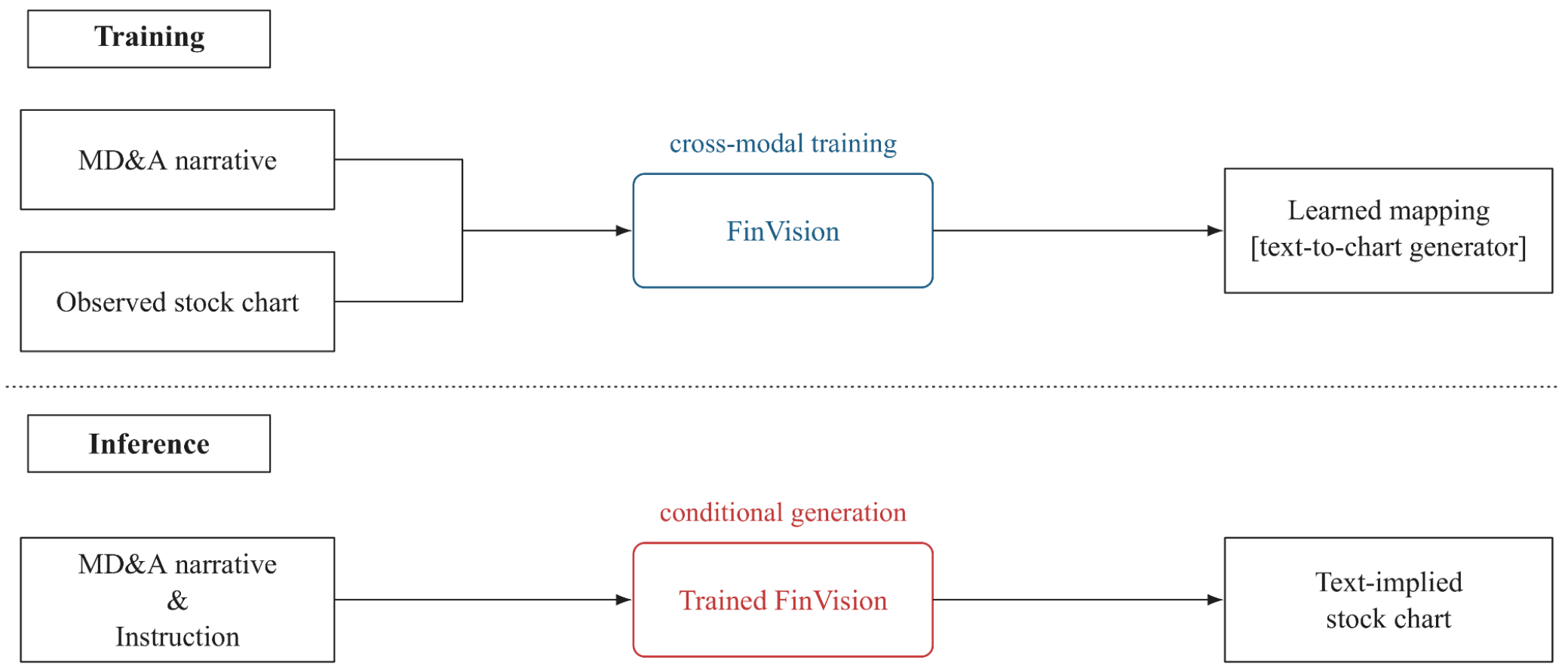

**Figure 2. Overview of FinVision training and inference.** FinVision learns the mapping between MD&A narratives and observed stock charts during training and uses the learned mapping to generate text-implied stock charts from MD&A narratives during inference.

### 3.1 Representing Financial Narratives through Market Dynamics

We focus on the Management Discussion and Analysis (MD&A) section of 10-K filings as the financial narrative. It discusses the firm's financial condition and results of operations and explains important trends, events, and uncertainties affecting the business over the course of a fiscal year. To construct the corresponding visual representation, we use the firm's stock price chart over the same fiscal year. Stock prices evolve as market participants process and trade on information concerning the firm, and the resulting chart preserves the temporal structure of these market dynamics. In contrast to a single annual return, a stock chart captures how market conditions evolve throughout the year, including changes in direction, fluctuations, reversals, and volatility.

It is worth noting that we do not assume a causal relationship between the narrative and the realized stock-price path. The market may respond differently from what is implied by managerial disclosure, as stock prices reflect many other sources of information beyond the 10-K. Our rationale is instead that the two provide related views of the same firm and economic period: the MD&A describes the firm's experience through managerial communication, while the stock chart represents contemporaneous market dynamics.

Specifically, for firm $i$ and fiscal year $t$, we pair an MD&A for fiscal year $t$ with the firm's stock chart during fiscal year $t$:

$$X_{i,t}^{text} \rightarrow Y_{i,t}^{chart},$$

where $X_{i,t}^{text}$ denotes the MD&A of firm $i$ for fiscal year $t$, and $Y_{i,t}^{chart}$ denotes the firm's stock chart over the same fiscal year. Our objective is to learn the mapping from financial narratives to the visual market patterns that are typically associated with them across firms and years.

### 3.2 Stock Price Chart Representation

We represent market dynamics using the firm's stock price chart over the fiscal year. Following Jiang, Kelly, and Xiu (2023), we construct the chart from daily opening, high, low, and closing (OHLC) prices. This representation preserves both the temporal ordering of market movements and within-day price variation. For each trading day, the high and low prices are represented by the upper and lower endpoints of a vertical bar. The opening and closing prices are represented by short horizontal ticks on the left and right sides of the bar, respectively. The bars are arranged chronologically along the horizontal axis over the fiscal year.

**Table 1. Examples of Paired MD&A Text and Stock Charts.** Each row shows an excerpt from a firm's MD&A and its observed stock chart for the same fiscal year.

| Firm / fiscal year | MD&A excerpt | Observed stock chart |
|---|---|---|
| **Cepheid**<br>FY 2007 | "Total product sales increased 41% to $116.5 million in 2007 from $82.4 million in 2006. The increase in instrument sales was primarily due to an increase of $18.6 million of GeneXpert systems sales and $5.3 million Smart Cycler system sales." | 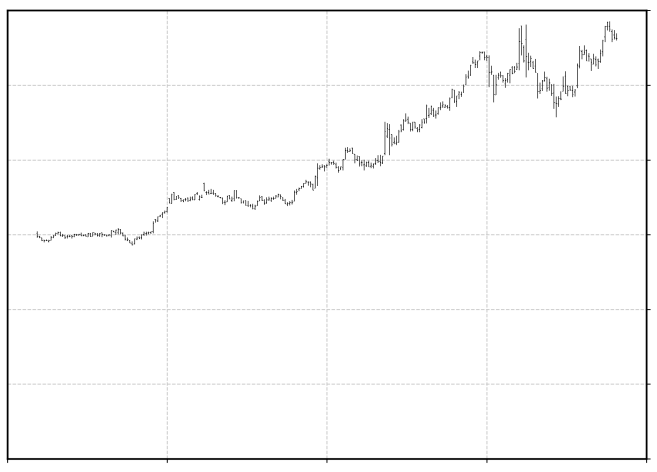 |
| **Westell Technologies**<br>FY 2009 | "In fiscal year 2009, consolidated revenue decreased by 22%. CNS equipment revenue decreased by 39%, OSPlant Systems equipment revenue increased 4% and ConferencePlus services revenue decreased by 16%." | 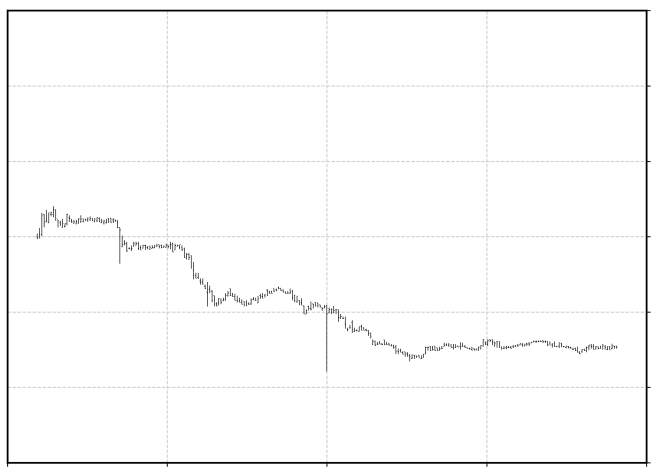 |

We standardize the charts to ensure that the model learns patterns of stock-price dynamics rather than firm-specific price levels or visual conventions. First, all charts are rendered in grayscale, removing color as a potential source of information. Second, stock prices are normalized within each firm-year chart to a common scale. The stock price on the first trading day is positioned at the midpoint of the vertical axis, and subsequent prices are represented relative to this initial level. This transformation removes differences in absolute price levels across firms while preserving relative price movements over the fiscal year. Table 1 provides an illustration of the stock-chart representation used in this study.

The stock chart captures several characteristics of market dynamics simultaneously. First, price movements relative to the initial level capture the overall direction and magnitude of change. Second, the sequence of daily observations preserves the trajectory of these changes over time. Third, the OHLC bars further capture variation in within-day price movements. Consequently, two firms with similar beginning-to-end returns can exhibit substantially different visual patterns when their price trajectories differ over the year. The stock chart therefore provides a structured representation of market dynamics beyond a single numerical outcome such as annual stock return.

Formally, let the daily market information for firm $i$ on trading day $d$ of fiscal year $t$ be

$$p_{i,t,d} = \left(O_{i,t,d}, H_{i,t,d}, L_{i,t,d}, C_{i,t,d}\right),$$

where $O$, $H$, $L$, and $C$ denote the opening, high, low, and closing prices, respectively. The ordered sequence

$$P_{i,t} = \left\{p_{i,t,1}, p_{i,t,2}, \ldots, p_{i,t,D_{i,t}}\right\}$$

contains the firm's daily OHLC observations over the fiscal year. We apply the normalization described above and render this sequence as a $512 \times 512$ grayscale image,

$$Y_{i,t}^{chart} = R(P_{i,t}),$$

where $R(\cdot)$ denotes the chart-rendering procedure. $Y_{i,t}^{chart}$ serves as the visual target paired with the firm's MD&A $X_{i,t}^{text}$.

### 3.3 FinVision Architecture

We now introduce FinVision, our domain-specific multimodal generative model for translating financial narratives into visual representations of contemporaneous market dynamics. As illustrated in Figure 3, FinVision consists of three main components: a text encoder that represents the information contained in an MD&A, an image autoencoder that represents stock price charts in a compressed latent space, and a Diffusion Transformer (DiT) that learns the cross-modal mapping between the textual and visual representations.

**Text Encoder.** We use Qwen2-VL-2B-Instruct (Wang et al., 2024) as the text encoder. The MD&A sections of 10-K filings are substantially longer and more specialized than typical prompts used in general-purpose text-to-image generation. Qwen2-VL processes the MD&A and produces a sequence of contextualized hidden representations for its textual content. We use the final hidden states as the textual representation conditioning the subsequent chart-generation process and apply RMS normalization to stabilize their scale. We keep the parameters of the text encoder fixed throughout FinVision training.

Formally, for the MD&A $X_{i,t}^{text}$, the text encoder $E_T(\cdot)$ produces the conditioning representation

$$c_{i,t} = E_T(X_{i,t}^{text}),$$

where $c_{i,t}$ contains the contextualized representations of the MD&A used to condition chart generation.

**Image Autoencoder.** We use a Deep Compression Autoencoder (DC-AE) to represent stock price charts in a lower-dimensional latent space (Chen et al., 2025). Given the stock chart $Y_{i,t}^{chart}$ defined in Section 3.2, the encoder component of the DC-AE maps the 512 × 512 chart into a compressed visual representation:

$$z_{i,t} = E_V(Y_{i,t}^{chart}),$$

where $E_V(\cdot)$ denotes the visual encoder and $z_{i,t}$ is the latent representation of the chart. The decoder $D_V(\cdot)$ performs the reverse transformation, converting a generated visual latent into a stock chart. We use a 32× compression ratio, which substantially reduces the dimensionality on which the generative model operates while preserving the visual structures of the stock chart. The image autoencoder is also kept fixed throughout training.

**Diffusion Transformer.** The DiT is the component through which FinVision learns the relationship between the MD&A representation $c_{i,t}$ and the stock-chart representation $z_{i,t}$. The DiT operates in the visual latent space and is conditioned on the textual representations produced by the text encoder. During training, it learns to recover the latent representation of an observed stock chart from a noisy version of that representation while conditioning on the corresponding MD&A. Across a large number of MD&A–chart pairs, this process allows the DiT to learn systematic associations between information expressed in financial narratives and visual patterns in stock-price dynamics.

We freeze both the text encoder and image autoencoder and train only the DiT, allowing model learning to focus on the cross-modal mapping from financial narratives to visual representations of market dynamics. During training, the observed stock chart is first encoded into a visual latent representation, to which random noise is added. Conditioned on the corresponding MD&A representation, the DiT learns to predict the added noise and recover the underlying chart latent. At inference, FinVision starts from a random visual latent and progressively denoises it under the guidance of the MD&A representation, producing a text-implied stock chart.

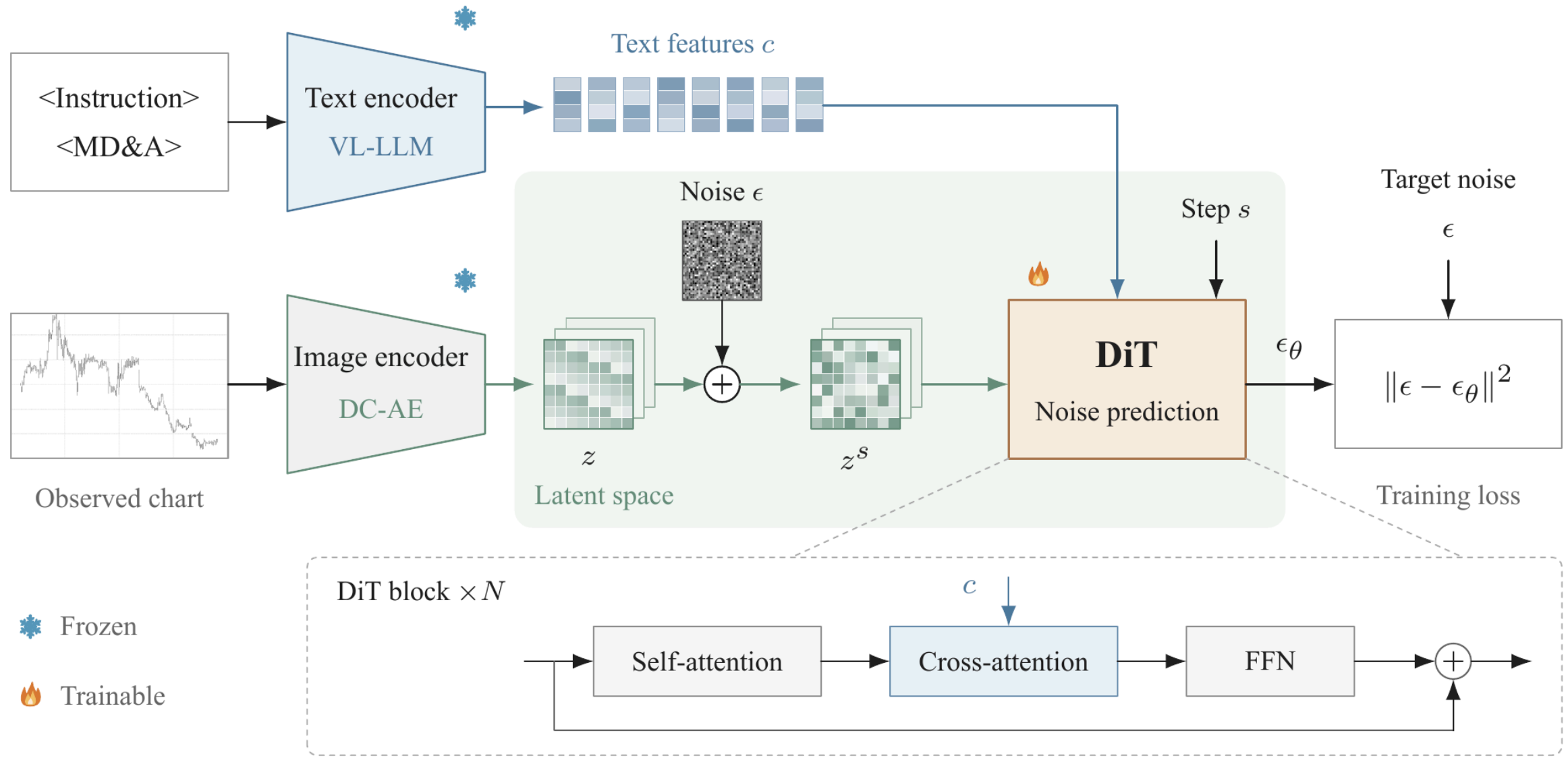

**Figure 3. FinVision training pipeline.** This figure shows how FinVision is trained. The frozen image encoder compresses an observed stock chart into a latent representation. Noise is added to this representation at a randomly sampled step. The Diffusion Transformer predicts the added noise from the text features, and the model is trained to minimize the prediction error.

### 3.4 FinVision Training

We train FinVision using 87,940 MD&A–stock chart pairs from annual reports filed between 1996 and 2020. Each training observation consists of an MD&A and the stock price chart for the same firm and fiscal year, $(X_{i,t}^{text}, Y_{i,t}^{chart})$.

For each training observation, the frozen text encoder first transforms the MD&A $X_{i,t}^{text}$ into the conditioning representation $c_{i,t}$. In parallel, the frozen image autoencoder transforms the corresponding stock chart $Y_{i,t}^{chart}$ into its latent representation $z_{i,t}$. FinVision then learns the conditional distribution of stock-chart representations given financial narratives.

Specifically, during each training step, noise is added to the latent representation of the observed stock chart at a randomly sampled diffusion timestep s, producing a noisy latent representation $z_{i,t}^{s}$. The Diffusion Transformer receives the noisy visual latent, the diffusion timestep, and the MD&A representation as inputs. It is trained to estimate the noise added to the observed chart representation conditional on the information contained in the MD&A.

The training objective is

$$L_{FinVision} = E_{s,z,\epsilon,c}\left[ \| \epsilon - \epsilon_{\theta}\left(z^{s}, s, c\right) \|^{2} \right] \qquad (1)$$

where $z$ denotes the latent representation of the observed stock chart, $z^{s}$ denotes its noisy representation at diffusion timestep $s$, $\epsilon$ denotes the noise added to the latent representation, $c$ denotes the textual conditioning representation of the corresponding MD&A, and $\epsilon_{\theta}(\cdot)$ denotes the noise estimated by the Diffusion Transformer with parameters $\theta$.

Intuitively, the training task requires FinVision to recover the visual structure of an observed stock chart while using the corresponding financial narrative as conditioning information. Repeating this task across MD&A–chart pairs allows the model to learn which textual descriptions of firm performance, business conditions, risks, and other economic developments are systematically associated with different stock-price patterns. Importantly, the model is not provided with researcher-defined mappings between textual features and market variables. These cross-modal associations are learned directly from the paired financial narratives and stock charts.

We additionally provide a chart-type instruction together with each MD&A to specify the desired visual representation. The model input follows the format

$$\langle Chart\ Type\ Instruction \rangle + \langle MD\&A\ Text \rangle.$$

In this study, we use the instruction: "*Please analyze the company's stock performance this year based on the following MD&A report and create a candlestick chart:*". This conditioning allows the same framework to generate the specified stock-chart representation while preserving the financial narrative as the primary information source.

We train the model on eight NVIDIA H800 GPUs using mixed-precision (FP16) training and gradient checkpointing. Optimization uses AdamW with a learning rate of $1 \times 10^{-4}$ and a batch size of 16 per GPU. The model is trained for one epoch, and stock charts are represented at a resolution of 512 × 512 pixels. We use a linear flow noise schedule with logit-normal timestep weighting during training.

### 3.5 Generating Text-Implied Stock Charts

After training, FinVision can generate a stock price chart from an MD&A. Given an MD&A $X_{i,t}^{text}$, the frozen text encoder first produces its textual representation,

$$c_{i,t} = E_T(X_{i,t}^{text}).$$

Chart generation begins from a randomly initialized noisy visual latent. Conditioned on $c_{i,t}$, the trained Diffusion Transformer progressively transforms this noisy latent into a visual representation associated with the information contained in the MD&A. The frozen image decoder then converts the generated latent representation into a 512 × 512 stock price chart. We denote the resulting chart as

$$\hat{Y}_{i,t}^{chart} = G_\theta(X_{i,t}^{text}),$$

where $G_\theta(\cdot)$ represents the complete FinVision generation process.

We refer to $\hat{Y}_{i,t}^{chart}$ as the **text-implied stock chart**. The term *text-implied* emphasizes that the generated chart is constructed solely from the information contained in the financial narrative and the cross-modal relationships learned during training. It is worth noting that the firm's actual stock chart $Y_{i,t}^{chart}$ is not provided to the model during inference. Thus, the generated chart represents the market dynamics that FinVision associates with a given MD&A based on patterns learned across in-sample text–chart pairs.

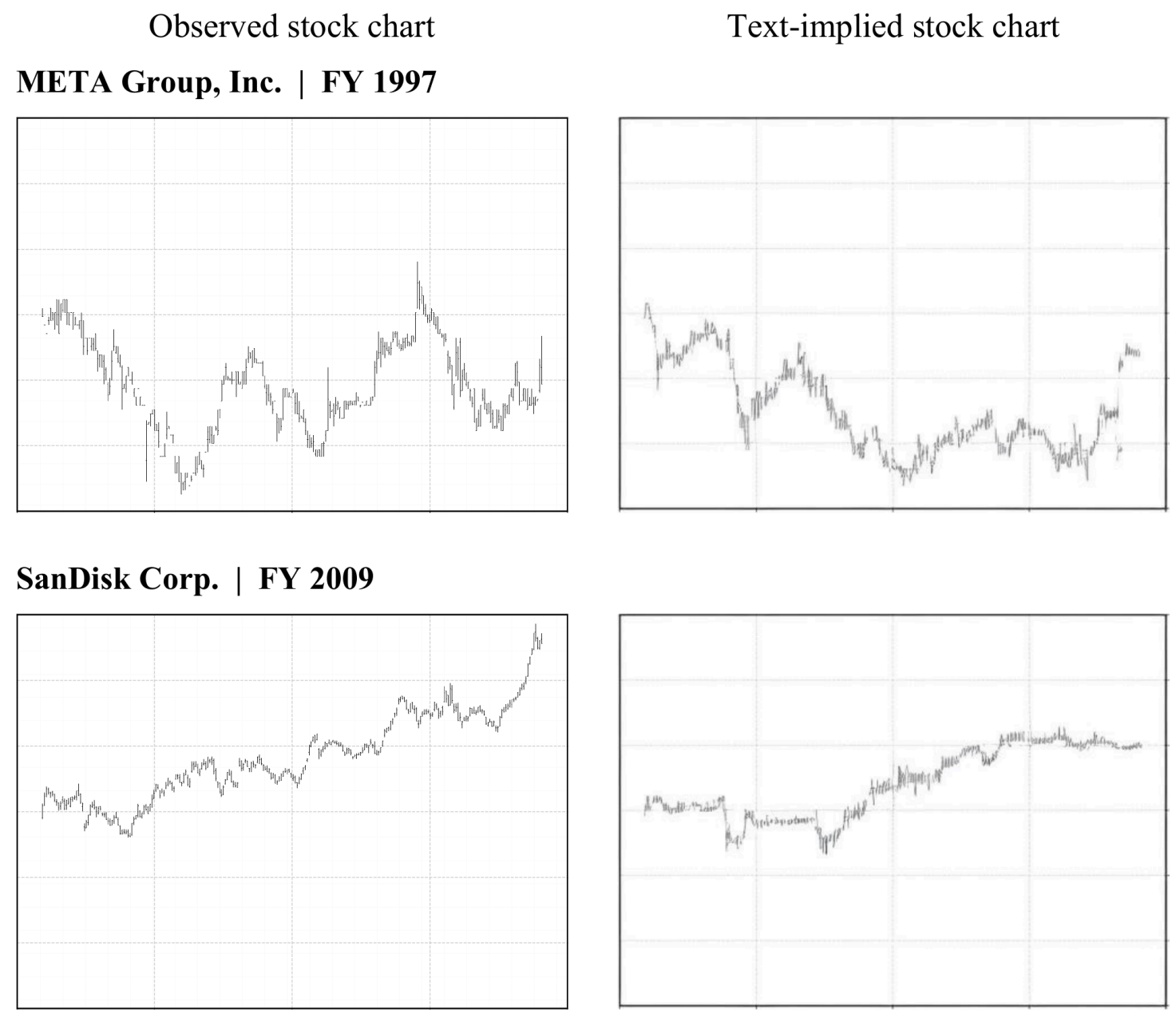

**Figure 4. Examples of text-implied stock charts.** Each row pairs an observed stock chart (left) with the text-implied stock chart generated by FinVision from the corresponding MD&A (right). The examples are META Group, Inc. in fiscal year 1997 and SanDisk Corp. in fiscal year 2009.

Figure 4 presents examples of text-implied stock charts generated by FinVision together with their corresponding observed stock charts. These examples illustrate that the model, after domain adaptation, is capable of generating complete price trajectories. Each generated chart simultaneously represents multiple characteristics of market dynamics, including the direction and magnitude of price movements, temporal variation in the trajectory, and variation in daily price ranges. The generated stock chart serves as the primary output representation of FinVision and forms the basis for the analyses that follow.

## 4. Visual Validation of FinVision

In this section, we first evaluate FinVision at the representation level by examining the visual similarity between the generated and observed stock charts. Specifically, we assess whether FinVision can generate stock-chart representations from financial narratives that resemble the contemporaneous charts observed in the market. In the next section, we turn to economic validation and examine whether these generated representations contain economically meaningful information.

We conduct this evaluation on an out-of-sample of 9,224 firm-year observations from 2023 and 2024, which do not overlap with the 1996–2020 training sample. For each MD&A, FinVision receives only the textual narrative and generates a text-implied stock chart, which we compare with the observed stock chart for the same firm and fiscal year. Because these firm-years are not used in model training, similarity between the generated and observed charts provides evidence that FinVision has learned a cross-modal relationship that generalizes beyond the training sample.

We compare FinVision against several baselines to turn corporate narratives into visual charts. The first baseline prompts an LLM directly: Qwen3-4B (Yang et al., 2025) reads the MD&A narrative and predicts a sequence of daily closing prices, which we then plot using Python. This represents a general-purpose prompting based text-to-number approach without task-specific training on paired financial narratives and stock charts. The second baseline uses general-purpose image generation models, specifically Stable Diffusion 3.5 Medium (Esser et al., 2024) and Qwen-Image (Wu et al., 2025). We select Stable Diffusion 3.5 Medium as a representative publicly available text-to-image model and Qwen-Image as a closer comparison because it uses the same Qwen-VL family for text conditioning. Both models receive the same financial narrative and generation instruction as FinVision, but neither is adapted to the financial narrative-to-chart task. FinVision differs from these baselines in that it is explicitly trained on paired 10-K MD&A narratives and contemporaneous stock charts.

Furthermore, to isolate the value of this financial-domain adaptation from the underlying architecture, we further evaluate a variant with the same FinVision architecture but trained on general-domain web images rather than paired financial narratives and stock charts (Wang et al., 2023). We denote this as FinVision-Web-Images.

We evaluate the text-implied stock chart similarity using five complementary image-based metrics, focusing on both fine-grained detail and overall appearance.

1. Pixel-level accuracy (PSNR): Peak Signal-to-Noise Ratio measures point-by-point accuracy on a decibel (dB) scale (Wang and Bovik, 2009). Intuitively, it measures how cleanly the generated chart reproduces the exact pixel coordinates of the real chart. Higher values mean less visual noise and closer point-by-point alignment.
2. Geometric structure (SSIM): The Structural Similarity Index (Wang et al., 2004) scores whether the overall shape, slope, and visual layout match the actual chart. SSIM ranges from 0 to 1, with higher scores indicating that the generated chart accurately follows the trajectory and turning points of the actual market trend.
3. Perceptual realism (LPIPS and DreamSim): Mathematical pixel comparisons can penalize minor coordinate shifts that human eyes barely notice. We therefore use Learned Perceptual Image Patch Similarity (Zhang et al., 2018) and DreamSim (Fu et al., 2023). Both metrics use neural networks trained on human visual judgments to evaluate whether the synthetic chart looks like a realistic stock chart. Lower scores mean smaller visual differences and higher realism.
4. Population-level distribution (CMMD): Beyond comparing individual charts pair by pair, we compute the CLIP Maximum Mean Discrepancy (Jayasumana et al., 2024) across the entire test set. This metric checks whether the full collection of 9,224 generated charts matches the statistical distribution of real stock charts. A lower

CMMD means the model produces realistic market variety rather than collapsing into repetitive patterns.

**Table 2. Out-of-sample evaluation on the 2023-2024 set (9,224 images per model).** PSNR/SSIM/LPIPS/DreamSim are per-image means against the paired ground truth; CMMD is a set-level score. Best in **bold**.

| | Pixel/structural | | Perceptual | | Distributional |
|---|---|---|---|---|---|
| Model | PSNR ↑ | SSIM ↑ | LPIPS ↓ | DreamSim↓ | CMMD ↓ |
| LLM (Qwen3-4B) | 19.58 | 0.775 | 0.403 | 0.341 | 1.885 |
| Stable Diffusion 3.5 Medium | 8.23 | 0.377 | 0.782 | 0.649 | 3.237 |
| Qwen-Image | 10.07 | 0.237 | 0.817 | 0.662 | 6.421 |
| FinVision (ours) | **25.98** | **0.905** | **0.202** | **0.102** | **1.490** |
| FinVision-Web-Images | 3.18 | 0.295 | 0.864 | 0.670 | 4.334 |

Table 2 reports the evaluation results. FinVision achieves the best score on all five metrics. The numerical baseline (Qwen3-4B with Python rendering) is the strongest external benchmark, but FinVision outperforms it across every dimension. FinVision raises PSNR from 19.58 to 25.98 dB and increases SSIM from 0.775 to 0.905. The advantage is even larger on perceptual metrics: LPIPS drops by 49.9% (from 0.403 to 0.202), DreamSim drops by 70.1% (from 0.341 to 0.102), and CMMD falls from 1.885 to 1.490. The relatively strong performance of Qwen3-4B may partly reflect its exposure to historical firm and market information during pretraining, which could introduce memorization or look-ahead effects when generating past price trajectories (He et al., 2026; Merchant and Levy, 2025; Gao et al., 2025). These results show that FinVision generates representations that more closely resemble the paired contemporaneous stock charts than the alternative approaches.

The general-purpose image generators perform poorly across all metrics. Stable Diffusion 3.5 and Qwen-Image achieve SSIM scores below 0.38 and high perceptual error (LPIPS above 0.78). Although these models excel at generating natural scenes and artistic images, they do not understand financial terminology or strict stock chart coordinates. When given a financial narrative, they often generate cluttered images with distorted axes and arbitrary lines. Furthermore, the comparison with the FinVision variant trained on general-domain web images highlights the importance of financial-domain adaptation. When we train the same FinVision architecture on general web images rather than paired financial narratives and stock charts, performance drops sharply. Its SSIM falls to 0.295 and LPIPS worsens to 0.864. This drop confirms that general image generation capabilities alone cannot produce reliable financial charts, and that the model must learn directly from paired disclosures and market chart data. In summary, the out-of-sample evaluation provides evidence that FinVision can generate visually meaningful stock-chart representations from financial narratives.

## 5. Economic Validation of FinVision

Having established the visual validity of FinVision, we next examine whether the generated representations contain economically meaningful information. For each firm-year, FinVision produces a text-implied stock chart based on the relationship learned from paired financial narratives and contemporaneous market charts. We compare this generated chart with the firm's observed stock chart over the same fiscal year and quantify the difference between the two. We refer to this difference as **narrative–market discrepancy**.

The intuition is straightforward. FinVision learns the stock-chart patterns that are typically associated with different financial narratives. The generated chart therefore reflects the market dynamics implied by the narrative based on this learned average relationship. The observed chart, in contrast, reflects the market dynamics that actually occurred during the same fiscal year. The difference between the two captures variation that is not represented by the average narrative-to-chart relationship learned by the model.

This variation can have several economic interpretations. One possibility is additional information: managers may provide explanations, private assessments, or forward-looking information that is not fully reflected in the historical market path (Li, 2010; Brown and Tucker, 2011). A narrative that differs from that path may therefore convey useful information. Another possibility is strategic disclosure: managers may selectively emphasize favorable developments, downplay unfavorable conditions, or frame performance in ways that shape investors' perceptions (Kothari, Shu, and Wysocki, 2009; Huang, Teoh, and Zhang, 2014). In this paper, we do not attempt to distinguish among these mechanisms. Instead, we evaluate the economic relevance of narrative–market discrepancy using outcomes commonly examined in the corporate disclosure literature. We first consider investor responses around the 10-K filing, which capture how the market reacts when the disclosure becomes public. We then examine subsequent market and operating outcomes to assess whether the discrepancy contains information about future firm performance.

### 5.1 Measuring Narrative–Market Discrepancy

For each MD&A $X_{i,t}^{text}$, FinVision generates a text-implied stock chart:

$$\hat{Y}_{i,t}^{chart} = G_{\theta}(X_{i,t}^{text}).$$

where $G_{\theta}(\cdot)$ denotes the FinVision generation process. Because the generated chart and the observed stock chart $Y_{i,t}^{chart}$ use the same visual representation, they can be directly compared.

We quantify the difference between the two using Learned Perceptual Image Patch Similarity (LPIPS; Zhang et al., 2018), which measures perceptual differences between images using features extracted from a pretrained neural network. The choice of LPIPS follows two considerations. First, the measure is constructed at the firm-year level, which rules out distribution-level metrics such as CMMD because they compare sets of images rather than individual generated–observed pairs. Second, it should reflect differences in the implied market pattern rather than in pixel-level rendering. LPIPS compares charts in a deep feature space calibrated to human visual judgments and is therefore less sensitive to small positional shifts than PSNR or SSIM. We use AlexNet as the feature-extraction backbone for LPIPS and define narrative–market discrepancy as

$$Discrepancy_{i,t}^{FinVision} = LPIPS\left(\hat{Y}_{i,t}^{chart}, Y_{i,t}^{chart}\right).$$

A larger value indicates a greater difference between the market dynamics implied by the financial narrative and those observed during the same fiscal year, while a smaller value indicates greater similarity between the two representations. Figures 5 and 6 provide examples of firm-years with relatively high and low discrepancy.

We also construct two benchmark discrepancy measures using the text-to-image models evaluated in Table 2. This allows us to assess whether the economic associations of narrative–market discrepancy arise from generic image generation or depend on financial-domain adaptation. The first uses Qwen-Image, which uses the same Qwen-VL family for text conditioning. The second is FinVision-Web-Images, which uses the same FinVision architecture trained on general-domain web images. For each benchmark, we generate a chart from the firm's MD&A and compare it with the observed chart using LPIPS.

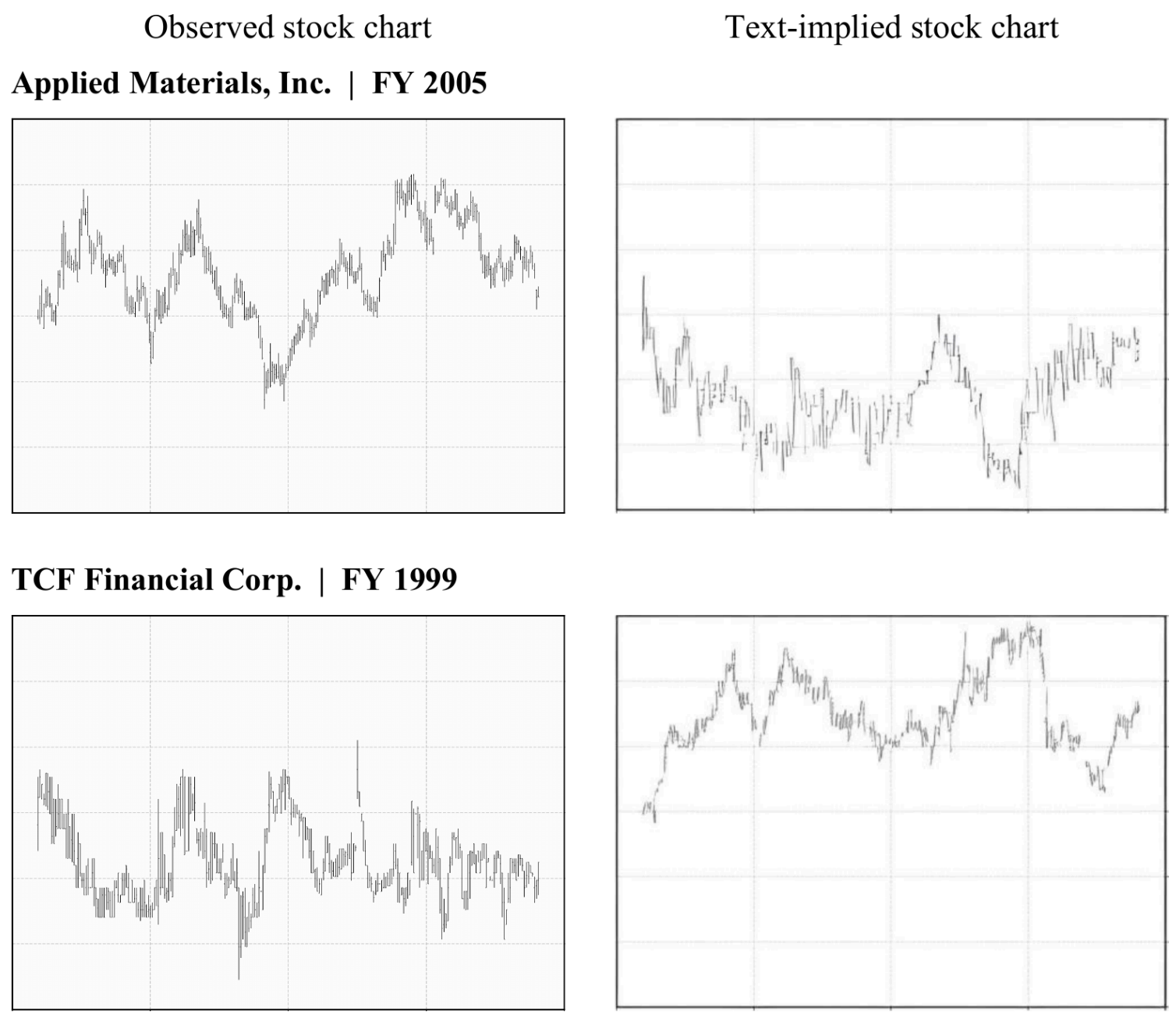

**Figure 5**. **Examples of high narrative–market discrepancy**. The figure presents two pairs of text-implied and observed stock charts with relatively large LPIPS-based discrepancy.

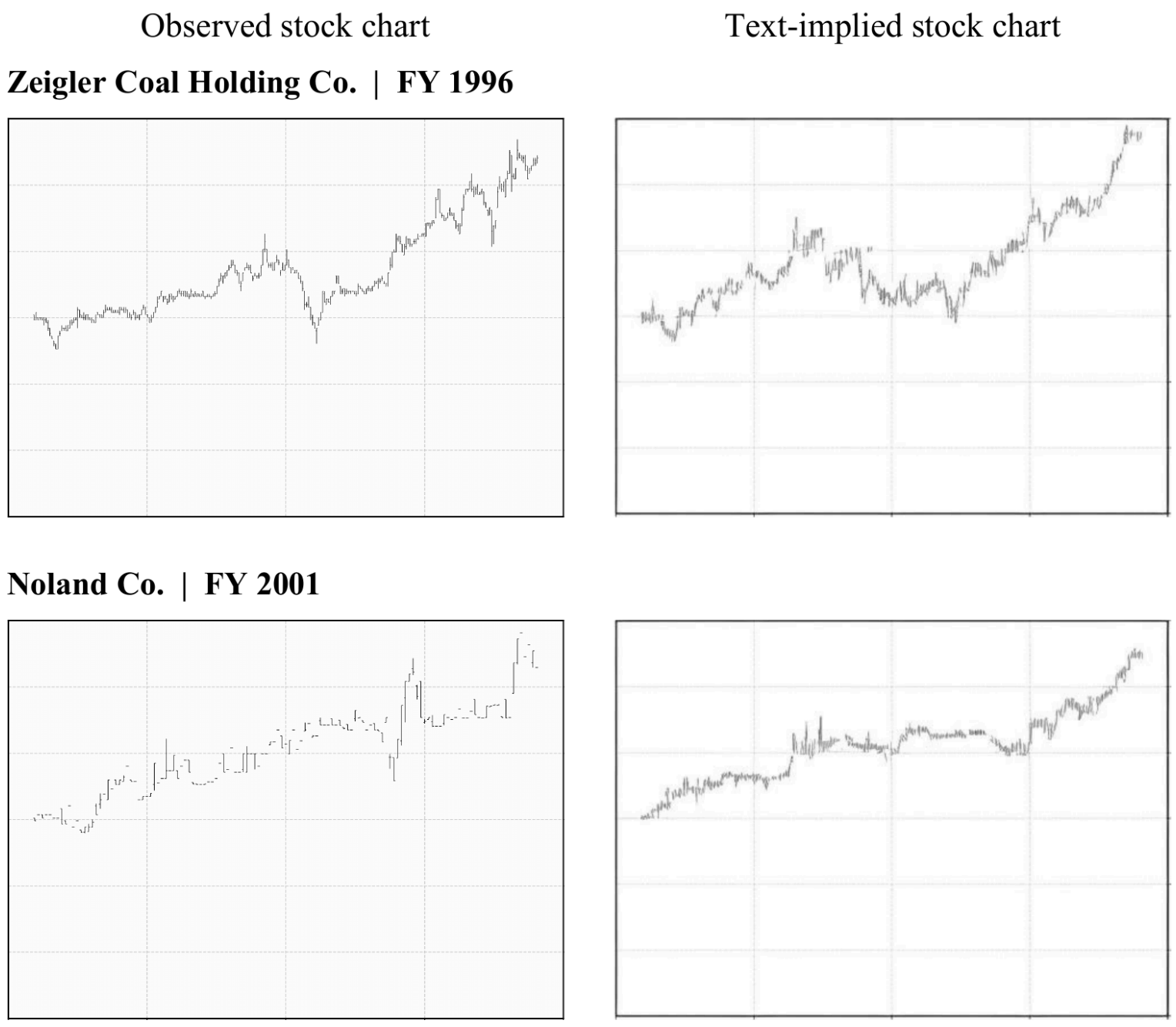

**Figure 6. Examples of low narrative–market discrepancy.** The figure presents two pairs of text-implied and observed stock charts with relatively small LPIPS-based discrepancy.

## 5.2 Narrative–Market Discrepancy and Investor Responses

Our first analysis examines whether narrative–market discrepancy is associated with investor responses around the subsequent 10-K filing. This setting provides a useful temporal separation between the construction of the discrepancy measure and the response being examined. The generated and observed charts correspond to the fiscal year described in the MD&A, whereas investor responses are measured around the later filing date.

If the difference between the text-implied and observed market representations captures economically meaningful variation, it should be systematically related to how investors respond when the disclosure becomes publicly available. We estimate the following regression:

$$MarketReaction_{i,t} = \alpha + \beta Discrepancy_{i,t} + \gamma Controls_{i,t} + FirmFE + YearFE + \varepsilon_{i,t}.$$

We use two measures of market response: the absolute value of the three-day market-adjusted return centered on the 10-K filing date ($|CAR^{10K}|$) and the cumulative three-day abnormal trading volume around the filing date ($AbVolume^{10K}$). Following Brown and Tucker (2011), we control for firm size, filing delay, additional financial information, and the absolute value of unexpected earnings. All Discrepancy measures are standardized to have a mean of zero and a standard deviation of one. All regressions include firm and year fixed effects, and standard errors are clustered at the firm level. Variable definitions are in Appendix A. Descriptive statistics are reported in Appendix B.

Table 3, Panel A reports the baseline results. In Column (1), the coefficient on FinVision-based *Discrepancy* is negative and statistically significant for $|CAR^{10K}|$. A one-standard-deviation increase in narrative–market discrepancy is associated with a 10.2-basis-point reduction in the magnitude of the cumulative abnormal return around the filing date. Column (2) similarly shows a negative association between FinVision-based *Discrepancy* and abnormal trading volume. These results indicate that larger differences between the text-implied and observed market representations are associated with weaker subsequent investor responses when the 10-K is released.

Table 3, Panel B reports the joint regressions. Columns (1) and (2) include Qwen-Image-based discrepancy alongside the FinVision measure. The FinVision coefficients remain negative and significant for absolute returns and abnormal volume, while the Qwen-Image coefficients are not significant. Columns (3) and (4) replace Qwen-Image with FinVision-Web-Images. The FinVision coefficients remain close to the baseline estimates, whereas the benchmark coefficients are again not significant. The coefficient-difference tests provide a direct comparison within each joint specification. For absolute returns, the FinVision coefficient is more negative than the Qwen-Image coefficient ($p = 0.015$) and the coefficient of FinVision-Web-Images ($p < 0.01$). For abnormal volume, the difference is significant at the 10% level relative to Qwen-Image ($p = 0.082$), but is not significant relative to the variant FinVision-Web-Images ($p = 0.178$). In sum, the joint regressions show that the FinVision-based measure contains incremental information about investor responses beyond the discrepancy measures produced by either baseline.

One possible interpretation is that greater narrative–market discrepancy reflects a disclosure that is less consistent with information already reflected in the market. Investors may therefore place less weight on such disclosures when they are released, resulting in a smaller market response. This interpretation is consistent with prior research showing that market reactions to managerial disclosures depend on their informativeness and perceived credibility (Brown and Tucker, 2011; Yang, 2012). Additional cross-sectional analyses reported in Appendix C show that the negative association is stronger when firms perform poorly and weaker when external monitoring is stronger.

**Table 3. Narrative-Market Discrepancy and Investor Responses.** This table reports OLS regressions relating narrative–market discrepancy to investor responses around the 10-K filing date. Panel A reports baseline regressions using discrepancy constructed with FinVision. Panel B reports joint regressions including FinVision-based discrepancy and a benchmark discrepancy measure. Columns (1)–(2) use Qwen-Image as the benchmark, and Columns (3)–(4) use FinVision-Web-Images. The final two rows of Panel B report the difference between the FinVision and benchmark discrepancy coefficients and the p-value for the null hypothesis that the coefficients are equal. All *Discrepancy* measures are normalized to have a mean of zero and standard deviation of one. All regressions include year and firm fixed effects. t-stats based on standard errors clustered by firm are reported in parentheses below the coefficients. *, **, and *** indicate two-tailed significance at the 10%, 5%, and 1% levels, respectively. Variable definitions are in Appendix A.

**Panel A: Baseline Results Using FinVision**

| | (1) | (2) |
|---|---|---|
| Dependent Variables = | $\|CAR^{10k}\|$ | $AbVolume^{10k}$ |
| $Discrepancy^{FinVision}$ | -0.102*** | -0.024* |
| | (-4.51) | (-1.95) |
| *Size* | -0.408*** | 0.351*** |
| | (-6.78) | (7.37) |
| *FileLate* | 0.999*** | 0.126* |
| | (7.11) | (1.86) |
| *Newitems* | 1.556** | -0.209 |
| | (2.03) | (-0.43) |
| \|*Unexpected Earnings*\| | 0.412*** | 0.017 |
| | (6.39) | (0.48) |
| Intercept | -1.846 | -1.478 |
| | (-0.42) | (-0.54) |
| Year FE | Y | Y |
| Firm FE | Y | Y |
| Observations | 49,332 | 47,716 |
| Adjusted $R^2$ | 0.2327 | 0.3639 |

**Panel B: Joint Regressions with Benchmark Discrepancy Measures**

| | (1) | (2) | (3) | (4) |
|---|---|---|---|---|
| Dependent Variables = | $\|CAR^{10k}\|$ | $AbVolume^{10k}$ | $\|CAR^{10k}\|$ | $AbVolume^{10k}$ |
| $Discrepancy^{FinVision}$ | -0.101*** | -0.025** | -0.102*** | -0.024* |
| | (-4.47) | (-1.98) | (-4.50) | (-1.95) |
| $Discrepancy^{Qwen\text{-}Image}$ | -0.017 | 0.009 | | |
| | (-0.68) | (0.66) | | |
| $Discrepancy^{Web\text{-}Images}$ | | | 0.015 | -0.001 |
| | | | (0.68) | (-0.11) |
| *Size* | -0.407*** | 0.350*** | -0.408*** | 0.351*** |
| | (-6.78) | (7.36) | (-6.78) | (7.37) |
| *FileLate* | 1.000*** | 0.125* | 1.000*** | 0.126* |
| | (7.11) | (1.85) | (7.11) | (1.86) |
| *Newitems* | 1.556** | -0.208 | 1.553** | -0.208 |
| | (2.03) | (-0.43) | (2.03) | (-0.43) |
| \|*Unexpected Earnings*\| | 0.412*** | 0.017 | 0.412*** | 0.017 |
| | (6.39) | (0.48) | (6.39) | (0.48) |
| Intercept | -1.849 | -1.478 | -1.832 | -1.479 |
| | (-0.42) | (-0.54) | (-0.42) | (-0.54) |
| Year FE | Y | Y | Y | Y |
| Firm FE | Y | Y | Y | Y |
| Observations | 49,332 | 47,716 | 49,332 | 47,716 |
| Adjusted $R^2$ | 0.2327 | 0.3639 | 0.2327 | 0.3639 |
| $\beta^{FinVision}$-$\beta^{Benchmark}$ | -0.084 | -0.034 | -0.117 | -0.023 |
| p-value | 0.015 | 0.082 | <0.01 | 0.178 |

### 5.3 Narrative–Market Discrepancy and Future Firm Outcomes

We next examine whether narrative–market discrepancy contains information about subsequent firm outcomes. This analysis provides a second form of economic validation by testing whether the variation captured by the difference between generated and observed charts is associated with future market and operating performance beyond conventional firm characteristics and prior performance.

We examine three outcomes in fiscal year $t + 1$: buy-and-hold abnormal stock returns ($BHAR$), cumulative abnormal returns around subsequent earnings announcements ( $CAR^{EA}$), and changes in return on assets ($\Delta ROA$). We estimate:

$$Outcome_{i,t+1} = \alpha + \beta Discrepancy_{i,t} + \gamma Controls_{i,t} + FirmFE + YearFE + \varepsilon_{i,t+1}.$$

The regressions control for prior market and operating performance as well as turnover, firm size, market-to-book ratio, leverage, abnormal accruals, gross profit, sales growth, analyst following, and institutional ownership. All regressions include firm and year fixed effects, and standard errors are clustered at the firm level.

Table 4, Panel A reports the baseline results. Greater FinVision-based discrepancy is consistently associated with weaker subsequent outcomes. A one-standard-deviation increase in *Discrepancy* is associated with a 1.4-percentage-point reduction in buy-and-hold abnormal returns in the subsequent year. FinVision-based *Discrepancy* is also negatively associated with abnormal returns around subsequent earnings announcements and with subsequent changes in return on assets.

We then include each baseline discrepancy measure alongside FinVision-based discrepancy and report results in Table 4, Panel B. Columns (1)–(3) report the joint regressions with Qwen-Image. FinVision-based discrepancy remains negatively associated with subsequent $BHAR$ and $\Delta ROA$, whereas the corresponding Qwen-Image coefficients are not significant. The FinVision coefficients are more negative than the Qwen-Image coefficients for both outcomes, with coefficient-difference p-values of 0.001 and 0.041, respectively. Relative to Qwen-Image, FinVision's incremental associations extend to both stock returns and changes in profitability, whereas Qwen-Image's significant association is confined to earnings-announcement returns. Columns (4)–(6) report the joint regressions with FinVision-Web-Images. The FinVision coefficients remain negative and significant for all three outcomes, with magnitudes close to their baseline estimates. None of the benchmark discrepancy coefficients is significant. This pattern supports the broader economic information content of the FinVision-based measure and highlights the value of financial-domain training.

These findings show that the FinVision-based discrepancy measure captures economically meaningful variation that is related to subsequent firm performance. One possible interpretation is that, larger discrepancy reflects unfavorable conditions that are not fully conveyed in the narrative and become more apparent in later market and operating outcomes. This interpretation is consistent with prior research on delayed or selective disclosure of bad news (Kothari, Shu, and Wysocki, 2009; Doyle, Jennings, and Soliman, 2013; Huang, Teoh, and Zhang, 2014; Bao, Kim, Mian, and Su, 2019). Together with the weaker investor responses around the 10-K filing, these results indicate that narrative–market discrepancy captures economically relevant divergence between managerial communication and market experience. More broadly, these findings show that FinVision provides a new way to analyze corporate disclosure by translating narrative information into a visual modality that can be directly compared with observed market dynamics.

**Table 4. Narrative-Market Discrepancy and Future Performance.** This table reports OLS regressions relating narrative–market discrepancy to subsequent firm outcomes. Panel A reports baseline regressions using discrepancy constructed with FinVision. Panel B reports joint regressions including FinVision-based discrepancy and a benchmark discrepancy measure. Columns (1)–(3) use Qwen-Image as the benchmark, and Columns (4)–(6) use FinVision-Web-Images. The final two rows of Panel B report the difference between the FinVision and benchmark discrepancy coefficients and the p-value for the null hypothesis that the coefficients are equal. All *Discrepancy* measures are normalized to have a mean of zero and standard deviation of one. All regressions include year and firm fixed effects. t-stats based on standard errors clustered by firm are reported in parentheses below the coefficients. *, **, and *** indicate two-tailed significance at the 10%, 5%, and 1% levels, respectively. Variable definitions are in Appendix A.

**Panel A: Baseline Results Using FinVision**

| | (1) | (2) | (3) |
|---|---|---|---|
| | $BHAR_{t+1}$ | $CAR^{EA}_{t+1}$ | $\Delta ROA_{t+1}$ |
| $Discrepancy_t^{FinVision}$ | -0.014*** | -0.086* | -0.002** |
| | (-4.67) | (-1.74) | (-2.07) |
| $BHAR_t$ | -0.177*** | -0.111 | 0.022*** |
| | (-26.14) | (-1.11) | (9.57) |
| $CAR^{EA}_t$ | 0.011*** | -0.134*** | 0.001*** |
| | (27.38) | (-19.05) | (11.00) |
| $\Delta ROA_t$ | -0.030 | -0.057 | -0.318*** |
| | (-1.53) | (-0.20) | (-24.97) |
| $BHAR_{t-1}$ | -0.096*** | | |
| | (-15.82) | | |
| $CAR^{EA}_{t-1}$ | | -0.078*** | |
| | | (-11.94) | |
| $\Delta ROA_{t-1}$ | | | -0.112*** |
| | | | (-11.69) |
| $Turnover_t$ | -0.019*** | -0.110** | -0.001 |
| | (-5.15) | (-1.97) | (-0.80) |
| $Size_t$ | -0.164*** | -0.679*** | -0.018*** |
| | (-17.25) | (-4.83) | (-5.04) |
| $MtoB_t$ | -0.002** | 0.002 | 0.001** |
| | (-2.10) | (0.14) | (2.46) |
| $Leverage_t$ | 0.167*** | 0.958** | 0.099*** |
| | (5.09) | (2.05) | (7.67) |
| $\lvert Abnormal\ Accrual \rvert_t$ | 0.010 | 0.090 | 0.036*** |
| | (0.46) | (0.25) | (3.48) |
| $Gross\ Profit_t$ | -0.000 | -0.005 | -0.001 |
| | (-0.07) | (-0.18) | (-0.98) |
| *Sales Growth $_t$* | -0.028* | -0.508** | -0.016*** |
| | (-1.83) | (-2.06) | (-2.87) |
| *Analyst Following $_t$* | -0.147*** | -0.359** | -0.007*** |
| | (-14.37) | (-2.23) | (-2.67) |
| *Institutional Ownership $_t$* | -0.115*** | 0.463 | -0.033*** |
| | (-4.56) | (1.15) | (-5.20) |
| Intercept | 1.490*** | 5.067*** | 0.125*** |
| | (24.68) | (5.80) | (5.91) |
| Year FE | Y | Y | Y |
| Firm FE | Y | Y | Y |
| Observations | 38,434 | 38,434 | 38,434 |
| Adjusted $R^2$ | 0.1827 | 0.0613 | 0.1336 |

**Panel B: Joint Regressions with Benchmark Discrepancy Measures**

| | (1) | (2) | (3) | (4) | (5) | (6) |
|---|---|---|---|---|---|---|
| | $BHAR_{t+1}$ | $CAR^{EA}_{t+1}$ | $\Delta ROA_{t+1}$ | $BHAR_{t+1}$ | $CAR^{EA}_{t+1}$ | $\Delta ROA_{t+1}$ |
| $Discrepancy_t^{FinVision}$ | -0.014*** | -0.079 | -0.002** | -0.014*** | -0.086* | -0.002** |
| | (-4.68) | (-1.61) | (-2.12) | (-4.65) | (-1.75) | (-2.06) |
| $Discrepancy_t^{Qwen\text{-}Image}$ | 0.001 | -0.122** | 0.001 | | | |
| | (0.34) | (-2.25) | (0.89) | | | |
| $Discrepancy_t^{Web\text{-}Images}$ | | | | 0.002 | -0.017 | 0.000 |
| | | | | (0.71) | (-0.33) | (0.30) |
| *Controls $_t$* | Y | Y | Y | Y | Y | Y |
| Year FE | Y | Y | Y | Y | Y | Y |
| Firm FE | Y | Y | Y | Y | Y | Y |
| Observations | 38,434 | 38,434 | 38,434 | 38,434 | 38,434 | 38,434 |
| Adjusted $R^2$ | 0.1826 | 0.0614 | 0.1336 | 0.1826 | 0.0613 | 0.1336 |
| $\beta^{FinVision}$-$\beta^{Benchmark}$ | -0.015 | 0.043 | -0.003 | -0.016 | -0.069 | -0.002 |
| p-value | 0.001 | 0.561 | 0.041 | <0.001 | 0.331 | 0.101 |

## 6. Conclusion

Financial narratives provide important information about how firms evolve over time. This study introduces FinVision, a multimodal generative AI method that translates financial narratives into structured visual representations of market dynamics. Trained on paired MD&A disclosures and contemporaneous stock price charts, FinVision generates a text-implied stock chart that reflects the market patterns typically associated with a narrative.

The central contribution of this study is methodological. Whereas traditional approaches transform financial narratives into textual constructs, numerical variables, or latent vectors, FinVision maps them directly into a visual modality, extending financial text analysis from text-to-variable toward text-to-visual representation. We validate this approach in two ways. Out-of-sample tests show that FinVision generates charts that resemble contemporaneous market patterns, while differences between generated and observed charts are systematically associated with investor responses and subsequent firm outcomes. Together, these results show that the generated visual representations are both visually and economically meaningful.

More broadly, our study illustrates how generative multimodal AI can use one modality to enrich the representation of another (Shin et al., 2020). This perspective may extend beyond finance. For example, in healthcare, clinical notes can be paired with medical images, vital signs, or other physiological signals to support cross-modal representation learning and downstream predictive analytics (Meyer et al., 2014; Morid and Sheng, 2025). Similar domain-adapted approaches may be useful whenever different data modalities can be paired and provide complementary views of the same underlying phenomenon.

This study has several limitations that can be further addressed. First, FinVision is developed using MD&A disclosures and stock price charts, and future work could examine other forms of business communication and other visual charts. Second, the current framework focuses on stock-price dynamics. Future research could incorporate other market signals, such as trading volume or volatility, or consider representations beyond financial markets. Third, the mapping learned by FinVision remains difficult to interpret at the level of individual textual inputs. Developing methods to identify which parts of a narrative drive specific features of the generated representation would improve transparency and deepen our understanding of how multimodal models encode dynamic business information.

**Online Appendix A: Variable Definitions**

| Variable Name | Definition |
|---|---|
| *Discrepancy* $^{Model}$ | It measures the difference between the generated and actual stock price images. The superscript indicates the model used to construct the measure: FinVision, Qwen-Image, or FinVision-Web-Images. Each model-specific measure is standardized to have a mean of zero and a standard deviation of one before inclusion in the regression. |
| $\|CAR^{10k}\|$ | The absolute value of the three-day market-adjusted return centered on the 10-K filing date, multiplied by 100. |
| $AbVolume^{10k}$ | The cumulative three-day abnormal share turnover centered on the 10-K filing date, multiplied by 100. Abnormal share turnover equals daily share turnover, subtracting the market turnover. |
| *Size* | The natural logarithm of the total assets at the end of a fiscal year. |
| *FileLate* | An indicator variable that equals one if the 10-K filing date is at least 90 days after the fiscal year end and zero otherwise |
| *Newitems* | The natural logarithm of 1 plus the number of non-missing items on Compustat at the end of fiscal year. |
| *\|Unexpected Earnings\|* | The absolute value of unexpected earnings. Unexpected earnings equal actual EPS minus consensus analyst forecasts issued before the fiscal year end, divided by the stock price at the end of the fiscal year. |
| *BHAR* | The buy-and-hold abnormal return (market adjusted) during the fiscal year. |
| $CAR^{EA}$ | The three-day market-adjusted return centered on the earnings announcement date, multiplied by 100. |
| $\Delta ROA$ | ROA in year t minus ROA in year t-1. ROA equals income before extraordinary items of the fiscal year, scaled by total assets at the beginning of the year. |
| *Turnover* | Average monthly turnover, calculated as the monthly trading volume divided by the number of shares outstanding at the end of the month. |
| *MtoB* | Market value of equity plus book value of total liabilities scaled by book value of total assets at the end of the fiscal year. |
| *Leverage* | Sum of long-term debt and short-term debt of a firm-year, scaled by total assets at the end of the fiscal year. |
| *\|Abnormal Accrual\|* | The absolute difference between the firm's total accruals and normal accruals in a firm-year. For each of the 48 Fama-French industry-year, we estimate the modified Jones Model (Dechow, Sloan, and Sweeney, 1995) and use the estimated coefficients to calculate a firm-year's normal accruals. |
| *Gross Profit* | Sales minus cost of goods sold, scaled by sales. |
| *Sales Growth* | Sales in year t minus sales in year t-1, scaled by total assets at the beginning of the year. |
| *Analyst Following* | The natural logarithm of one plus the number of analysts following a firm. |
| *Institutional Ownership* | Percentage of institutional holdings as of the firm's most recent 13-F filing before the end of a fiscal year. |
| *Loss* | An indicator that equals one if net income before extraordinary items is negative, and zero otherwise. |
| *Neg BHAR* | An indicator that equals one if the buy-and-hold abnormal return (market adjusted) during the fiscal year is negative, and zero otherwise. |
| *High Default Risk* | An indicator that equals one if the z-score is below 1.8, and zero otherwise. This measure is derived from Altman's (1968) model, where higher z-scores signify a lower probability of bankruptcy. Z-score=1.2(working capital/total assets) + 1.4(retained earnings/total assets) + 3.3(earnings before interest and taxes/total assets) + 0.6(market value of equity/book value of total liabilities) + (sales/total assets). |
| *High* $\|AF_Error^{pre10k}\|$ | An indicator that equals one if Pre-10K forecast error is above the median, and zero otherwise. Pre-10K forecast error is measured as the absolute difference between actual earnings and the consensus analyst forecast for year t+1 (based on estimates issued in the 90-day window before the 10-K filing for year t). |

**Online Appendix B: Descriptive Statistics**

Table A1 provides descriptive statistics for the main variables used in the regressions. The raw FinVision-based *Discrepancy* measure has a mean of 0.206, a standard deviation of 0.035, and first and third quartiles of 0.182 and 0.231, respectively. The mean absolute three-day market-adjusted return around the 10-K filing is 4.570%. In the future-performance sample, mean fiscal-year buy-and-hold abnormal return is 0.050, the mean market-to-book measure is 3.176, and mean institutional ownership is 0.623. Mean analyst following, measured as the natural logarithm of one plus the number of analysts, is 1.897.

**Table A1. Descriptive Statistics.** This table reports the descriptive statistics of the main variables used in regressions. The variables *Discrepancy* is presented as raw values before normalization. Refer to Appendix A for detailed variable descriptions.

| Variable | N | Mean | SD | p25 | p50 | p75 |
|---|---|---|---|---|---|---|
| $Discrepancy^{FinVision}$ | 49,332 | 0.206 | 0.035 | 0.182 | 0.208 | 0.231 |
| $\|CAR^{10k}\|$ | 49,332 | 4.570 | 5.162 | 1.151 | 2.735 | 5.885 |
| $AbVolume^{10k}$ | 47,716 | -0.445 | 3.074 | -2.093 | -1.255 | 0.121 |
| *Size* | 49,332 | 6.346 | 1.858 | 4.965 | 6.270 | 7.614 |
| *FileLate* | 49,332 | 0.056 | 0.229 | 0.000 | 0.000 | 0.000 |
| *Newitems* | 49,332 | 5.707 | 0.137 | 5.613 | 5.756 | 5.808 |
| \|*Unexpected Earnings*\| | 49,332 | 0.162 | 0.846 | 0.001 | 0.003 | 0.014 |
| *BHAR* | 38,434 | 0.050 | 0.589 | -0.292 | -0.039 | 0.239 |
| $CAR^{EA}$ | 38,434 | 0.212 | 9.007 | -4.439 | 0.182 | 4.903 |
| $\Delta ROA$ | 38,434 | 0.006 | 0.198 | -0.036 | 0.000 | 0.032 |
| *Turnover* | 38,434 | 1.958 | 1.641 | 0.884 | 1.510 | 2.472 |
| *MtoB* | 38,434 | 3.176 | 5.490 | 1.243 | 2.106 | 3.721 |
| *Leverage* | 38,434 | 0.239 | 0.216 | 0.040 | 0.210 | 0.366 |
| \|*Abnormal Accrual*\| | 38,434 | 0.147 | 0.180 | 0.035 | 0.086 | 0.184 |
| *Gross Profit* | 38,434 | -0.072 | 3.106 | 0.218 | 0.361 | 0.542 |
| *Sales Growth* | 38,434 | 0.103 | 0.274 | -0.011 | 0.059 | 0.175 |
| *Analyst Following* | 38,434 | 1.897 | 0.763 | 1.386 | 1.946 | 2.485 |
| *Institutional Ownership* | 38,434 | 0.623 | 0.307 | 0.396 | 0.691 | 0.872 |

**Online Appendix C: Cross-Sectional Variation in Investor Responses**

We further examine whether the relation between narrative–market discrepancy and investor responses varies with firms' performance conditions and external information environments. Managers may have stronger incentives to selectively emphasize or withhold information when firm performance is poor (Li, 2008; Kothari, Shu, and Wysocki, 2009; Huang, Teoh, and Zhang, 2014). We therefore interact *Discrepancy* with three proxies for unfavorable conditions: *Loss*, *Neg BHAR*, and *High Default Risk*. As shown in Table A2, Panel A, all three interaction terms are negative and significant (*Loss* × *Discrepancy*: -0.115, t = -2.15; *Neg BHAR* × *Discrepancy*: -0.138, t = -3.14; *High Default Risk* × *Discrepancy*: -0.212, t = -3.74). These results indicate that the negative association between narrative–market discrepancy and investor responses is stronger when firms experience poorer performance or greater financial distress.

We also examine whether the relation varies with the external information environment. Stronger monitoring may constrain managerial discretion and provide investors with additional information for evaluating corporate disclosures (Chung, Firth, and Kim, 2002; Yu, 2008; Huang, Lehavy, Zang, and Zheng, 2018). Using *institutional ownership*, *analyst following*, and *pre-filing analyst forecast inaccuracy* as proxies, Table A2, Panel B shows that *High Institutional Ownership* × *Discrepancy* is positive and significant (0.091, t = 1.96), while *High Analyst Following* × *Discrepancy* is positive but insignificant. In contrast, *High |AF_Error Pre10K|* × *Discrepancy* is negative and significant (-0.151, t = -3.18). Overall, the association between narrative–market discrepancy and investor responses is stronger when firms face unfavorable conditions and weaker external monitoring.

**Table A2. Cross-sectional Variation in the Relation between MD&A Discrepancy and Market Reactions.** This table presents results from examining whether the relation between short-term market reactions and disclosure discrepancy varies with a set of firm characteristics. Panel A reports the variation in current operating performance and financial condition; Panel B reports the variation in investor and analyst characteristics. $Discrepancy^{FinVision}$ is normalized to have a mean of zero and standard deviation of one. All regressions include year and firm fixed effects. t-stats based on standard errors estimated clustered by firm are reported in parentheses below the coefficients. *, **, and *** indicate two-tailed significance at the 10%, 5%, and 1% levels, respectively. Variable definitions are in Appendix A.

**Panel A: Current performance and financial conditions**

| | (1) | (2) | (3) |
|---|---|---|---|
| Dependent Variable= | | $\lvert CAR^{10k} \rvert$ | |
| *Discrepancy* $^{FinVision}$ | -0.050** | -0.029 | -0.027 |
| | (-2.14) | (-0.97) | (-1.11) |
| *Loss* | 0.967*** | | |
| | (13.35) | | |
| *Loss* × *Discrepancy* $^{FinVision}$ | -0.115** | | |
| | (-2.15) | | |
| *Neg BHAR* | | 0.410*** | |
| | | (9.38) | |
| *Neg BHAR* × *Discrepancy* $^{FinVision}$ | | -0.138*** | |
| | | (-3.14) | |
| *High Default Risk* | | | 0.896*** |
| | | | (10.60) |
| *High Default Risk* × *Discrepancy* $^{FinVision}$ | | | -0.212*** |
| | | | (-3.74) |
| Controls | Y | Y | Y |
| Year FE | Y | Y | Y |
| Firm FE | Y | Y | Y |
| Observations | 49,332 | 49,332 | 49,332 |
| Adjusted $R^2$ | 0.2370 | 0.2343 | 0.2360 |
| *Discrepancy* + Interaction term | -0.165*** | -0.167*** | -0.239*** |
| p-value | <0.01 | <0.01 | <0.01 |

**Panel B: Investor and analyst characteristics**

| | (1) | (2) | (3) |
|---|---|---|---|
| Dependent Variable= | | $\lvert CAR^{10k} \rvert$ | |
| *Discrepancy* $^{FinVision}$ | -0.149*** | -0.140*** | -0.008 |
| | (-4.12) | (-3.74) | (-0.29) |
| *High Institutional Ownership* | -0.185** | | |
| | (-2.25) | | |
| *High Institutional Ownership* × *Discrepancy* $^{FinVision}$ | 0.091** | | |
| | (1.96) | | |
| *High Analyst Following* | | -0.161** | |
| | | (-2.19) | |
| *High Analyst Following* × *Discrepancy* $^{FinVision}$ | | 0.066 | |
| | | (1.44) | |
| *High* $\lvert AF_Error^{Pre10k} \rvert$ | | | 0.621*** |
| | | | (10.27) |
| *High* $\lvert AF_Error^{Pre10k} \rvert$ × *Discrepancy* $^{FinVision}$ | | | -0.151*** |
| | | | (-3.18) |
| Controls | Y | Y | Y |
| Year FE | Y | Y | Y |
| Firm FE | Y | Y | Y |
| Observations | 49,332 | 49,332 | 43,603 |
| Adjusted $R^2$ | 0.2329 | 0.2328 | 0.2358 |
| *Discrepancy* + Interaction term | -0.058** | -0.074*** | -0.159*** |
| p-value | 0.04 | <0.01 | <0.01 |